\documentclass[conference]{IEEEtran}
\usepackage[utf8]{inputenc}
\usepackage{amsmath}
\usepackage{graphicx}
\usepackage{amsfonts}
\usepackage{svg}

\usepackage[sorting=none]{biblatex} 
\usepackage{amsthm}
\theoremstyle{plain}
\newtheorem{theorem}{Proposition}

\title{Asymptotic Stability in Reservoir Computing}

\author{\IEEEauthorblockN{Jonathan Dong\IEEEauthorrefmark{1},
Erik Börve\IEEEauthorrefmark{1}, Mushegh Rafayelyan\IEEEauthorrefmark{2}, and
Michael Unser\IEEEauthorrefmark{1} \IEEEmembership{Fellow, IEEE}}
\IEEEauthorblockA{
\IEEEauthorrefmark{1}Biomedical Imaging Group, École Polytechnique Fédérale de Lausanne, Lausanne 1015, Switzerland \\
\IEEEauthorrefmark{2}PhotonicsAI lab, Department of Physics, Yerevan State University, Yerevan, Armenia}}

\begin{document}

\newcommand{\xx}{\mathbf{x}}
\newcommand{\yy}{\mathbf{y}}
\newcommand{\oo}{\mathbf{o}}
\newcommand{\ii}{\mathbf{i}}
\newcommand{\jj}{\mathbf{j}}
\newcommand{\uu}{\mathbf{u}}
\newcommand{\vv}{\mathbf{v}}
\newcommand{\ww}{\mathbf{w}}
\newcommand{\WW}{\mathbf{W}}
\newcommand{\GG}{\mathbf{G}}
\newcommand{\II}{\mathbf{I}}
\newcommand{\Geq}{G_\textrm{eq}}
\newcommand{\Gneq}{G_\textrm{neq}}
\newcommand{\gneq}{g_\textrm{neq}}
\newcommand{\LL}{\mathcal{L}}
\newcommand{\te}{^{(t)}}
\newcommand{\tee}{^{(t+1)}}
\newcommand{\rarcsin}{\frac{2}{\pi} \arcsin}

\maketitle

\begin{abstract}
    Reservoir Computing is a class of Recurrent Neural Networks with internal weights fixed at random. 
    Stability relates to the sensitivity of the network state to perturbations. It is an important property in Reservoir Computing as it directly impacts performance. 
    In practice, it is desirable to stay in a stable regime, where the effect of perturbations does not explode exponentially, but also close to the chaotic frontier where reservoir dynamics are rich.
    Open questions remain today regarding input regularization and discontinuous activation functions. 
    In this work, we use the recurrent kernel limit to draw new insights on stability in reservoir computing.
    This limit corresponds to large reservoir sizes, and it already becomes relevant for reservoirs with a few hundred neurons. 
    We obtain a quantitative characterization of the frontier between stability and chaos, which can greatly benefit hyperparameter tuning. 
    In a broader sense, our results contribute to understanding the complex dynamics of Recurrent Neural Networks. 
\end{abstract}


\section{Introduction}

Recurrent neural networks (RNN) represent a broad class of artificial neural networks. They present a temporal evolution with nonlinear internal dynamics and feedback loops, and are closer to biological neural networks compared to feedforward architectures. As such, their behavior is richer but harder to characterize. In particular, training RNNs remains a challenging problem and a topic of intense study \cite{pascanu2013difficulty}. For example, backpropagation typically exhibits exploding or vanishing gradients which limit our ability to train such networks.

To bypass this issue, reservoir computing (RC) fixes internal weights randomly and only adjusts the output weights of the network \cite{jaeger2001echo, verstraeten2007experimental}. This greatly facilitates the training process since one only has to learn a linear output model, with no problem of exploding or vanishing gradients. The philosophy behind RC is to use a large ensemble of randomly-connected neurons---the so-called \textit{reservoir}---to embed time-dependent information in a way such that a linear mapping to the desired output is possible. RC has been applied to different areas such as speech recognition \cite{larger2017high}, chaos cryptography \cite{antonik2018using}, and robot motor control \cite{lukovsevivcius2012reservoir}. It has been particularly promising for chaotic time series prediction \cite{pathak2018model}. 

The reservoir is a nonlinear dynamical system driven by an external input. Its properties need to be finely tuned to achieve optimal performance. On one hand, two distinct input patterns should lead to different reservoir states to distinguish them.
On the other hand, oversensitivity can also be detrimental as the reservoir can fall in a chaotic dynamical regime, where perturbations explode exponentially with time. It is thus important to have a stable reservoir, in which information about the current reservoir state and input vanishes exponentially at a controllable rate. This necessary condition has been stated in the founding paper \cite{jaeger2001echo} as the \textit{Echo-State Property}. The study of this transition between stable and chaotic regimes is enabled by the simplicity of RC, and we believe it may be relevant for other RNN architectures as well.

This stability property depends on hyperparameters of the reservoir---in particular, the standard deviation of the random and input weights. In practice, the best performance is typically obtained at the \textit{edge of chaos}, a stable regime close to the chaotic frontier where dynamics are richer \cite{lukovsevivcius2009reservoir}. Basic observations have been proposed to help with hyperparameter tuning. 
For instance, if the activation function is 1-Lipschitz continuous, then stability for any input is achieved when the largest singular value of the internal weight matrix is smaller than one \cite{wainrib2016local}. However, this bound is typically too conservative and a heuristic hyperparameter search is necessary for optimal performance.

This transition between stability and chaos raises several questions. Quantitative results are lacking for a broader class of activation functions like the rectified linear unit (ReLU, which is not differentiable at zero) or discontinuous activation functions. Moreover, it has been observed that the input regularizes the internal dynamics and allows for the use of spectral radii slightly larger than one \cite{lukovsevivcius2009reservoir}. A broader stability analysis would be beneficial since RC has been implemented in a large variety of settings \cite{appeltant2011information,larger2012photonic,hermans2015photonic,tanaka2019recent, rafayelyan2020large}; an example being physical implementations with binary activation functions \cite{dong2018scaling,dong2019optical}. 

Such questions are easier to tackle in the asymptotic limit, when the reservoir size is very large. In this large size limit, RC tends to a deterministic kernel that we iterate recurrently, called a recurrent kernel (RK) \cite{hermans2012recurrent,dong2020reservoir}. 
The powerful interpretation of RK enables a mean-field study of stability. However, it has only been applied to reproduce known results with continuous activations and no input \cite{hermans2012recurrent}. In another study, local Lyapunov exponents have been introduced to describe stability \cite{wainrib2016local}. They provide a quantitative analysis of the stability of RC in the presence of an input. However, this metric needs to be computed as we iterate the reservoir, making it computationally-demanding for applications such as hyperparameter search. Moreover, it cannot handle non-differentiable functions. 

In this paper, we show how this asymptotic limit enables us to quantitatively characterize stability in the presence of an input and with discontinuous activation functions. In particular, we exhibit two important properties of the activation function impacting stability---continuity and boundedness. 
This kernel limit greatly facilitates quantitative studies as stability boils down to analyzing fixed points iterations of deterministic functions. 

In Section 2, we give a basic description of RC and RK, along with a proper definition of stability in both cases. In Section 3, we then show how to apply this stability study to three representative examples: the error function activation (bounded and continuous), the sign function (bounded and discontinuous), and the Rectified Linear Unit or ReLU (unbounded and continous). 


\section{Theoretical Background}

\subsection{Reservoir Computing}

\subsubsection{Definition}

In the class of RC, we focus on the Echo-State Networks. These generic randomly-connected RNNs were first introduced by Jaeger \cite{jaeger2001echo} and are the most commonly-used ones in the field. A time-dependent input $\ii\te \in \mathbb{R}^d$ is fed into a reservoir of size $N$, yielding the following update equation for the reservoir state $\xx\te \in \mathbb{R}^N$:
\begin{equation}
    \xx\tee = \frac{1}{\sqrt{N}} f(\WW_r \xx\te + \WW_i \ii\te).
    \label{eq: reservoir update equation}
\end{equation}
Here, $f$ is an element-wise activation function, $\WW_r$ the internal reservoir weights, and $\WW_i$ the input weights. These weights are drawn from i.i.d. distributions $\mathcal{N}(0, \sigma_r^2)$ and $\mathcal{N}(0, \sigma_i^2)$, respectively. To emphasize the particularity of RC, these weights are fixed randomly and are not trained. Thus, iterating this update equation only depends on the input data. 

The algorithm generates an output $\oo\te$ using a linear model
\begin{equation}
    \oo\te = \WW_o \xx\te.
    \label{eq: linear output model}
\end{equation}
This training step consists of a simple linear regression. In RC, the complexity of the computation is performed during the nonlinear update described in Eq.~\eqref{eq: reservoir update equation}.


Current research directions include the physical implementation of RC on dedicated hardware. Thanks to the flexibility of RC, optical devices, dedicated electronics, and more exotic architectures have been proposed for small-footprint and fast computation. On top of that, recent software developments include Deep Reservoir Computing \cite{gallicchio2017deep,gallicchio2020deep}, in which a hierarchical architecture has been proposed to improve performance. 



\subsubsection{Stability}

\begin{figure*}
    \centering
    \includegraphics[width=\textwidth]{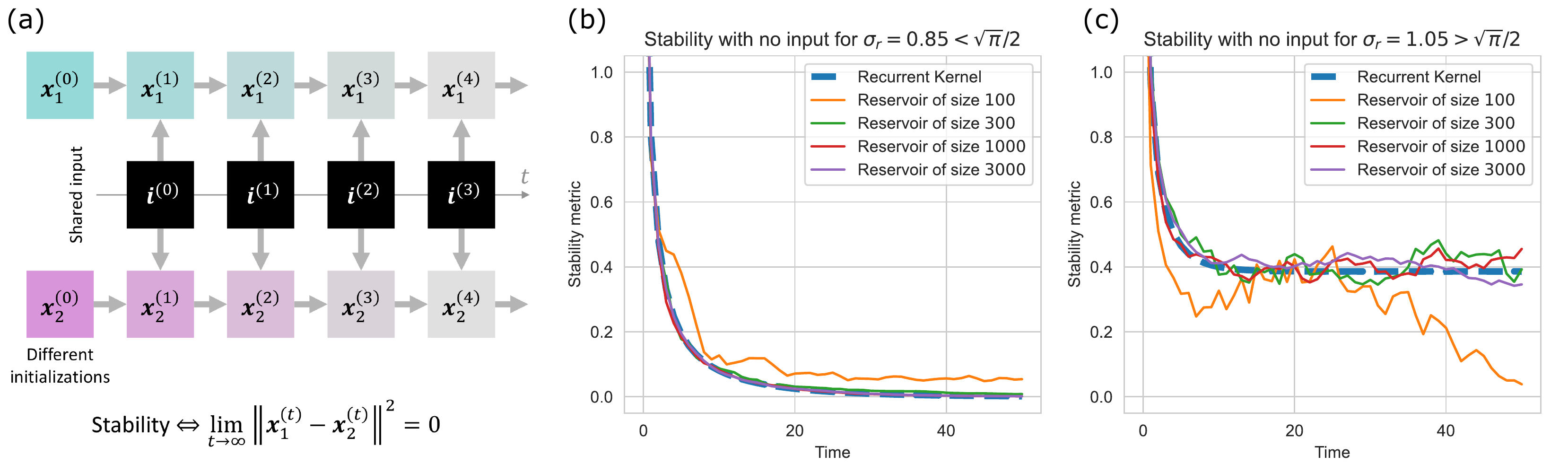}
    \caption{Principle and motivation for an asymptotic stability study. \textbf{(a) Scheme of a stability test.} To study stability in Reservoir Computing, two identical reservoirs are initialized differently and driven by the same input. The reservoir is stable if after a transient time, the reservoir state does not depend on this arbitrary initialization. We thus monitor the squared distance between the reservoir states through time. \textbf{(b-c) Asymptotic stability study in stable and chaotic cases.} Stability metric $L\te$ as a function of time $t$ for various reservoir sizes and for the corresponding Recurrent Kernel limit. The stable case corresponds to an erf activation function with $\sigma_r = 0.85$ and $\sigma_i = 0$. The chaotic case corresponds to the same previous parameters with the exception of $\sigma_r = 1.05$.}
    \label{fig1: stability}
\end{figure*}

As discussed previously, stability is a fundamental property to study in RC. This stability is typically characterized by the experiment depicted in Fig.~\ref{fig1: stability}a. We initialize two different reservoirs $\xx_1^{(0)}$, $\xx_2^{(0)}$ independently from i.i.d. Gaussian distributions. They share the same weights $\WW_r$ and $\WW_i$ and receive the same input $\ii\te$. At each time step, the input $\ii\te \in \mathbb{R}^d$ is randomly drawn from the unit sphere. 
We investigate whether these reservoirs converge towards a common trajectory. 

The stability metric we choose quantifies the distance between the two reservoir states as it evolves with time. It is given by
\begin{equation}
    L\te = \left\| \xx_1\te - \xx_2\te \right\|^2.
    \label{eq: metric reservoir computing}
\end{equation}
A reservoir configuration is called \textit{stable} for input $\ii\te$ if $\lim_{t\rightarrow\infty} L\te = 0$. Conversely, the Echo-State Property does not hold when $\lim_{t\rightarrow\infty} L\te > 0$. This is an indirect characterization of chaos; other behaviors such as several distinct fixed points would also be possible but both settings would not be suitable for a Reservoir Computing algorithm. 

The stability property depends on how the weights are set. The two parameters $\sigma_r^2$ and $\sigma_i^2$ will tune the transition between stability and chaos, by changing the variance of the random weights. Small internal weights exponentially damp the importance of previous reservoir states, while large internal weights tend to increase initial perturbations leading to a chaotic behavior. 

Stability has been characterized for Lipschitz-continuous functions for a random connectivity matrix \cite{wainrib2016local}. For example, since erf is $2 / \sqrt{\pi}$-Lipschitz, stability is ensured when $\sigma_r < \sqrt{\pi} / 2$. This result stands for any input and is optimal when there is zero input. Nevertheless, it is too conservative when an input is present.


\subsection{Recurrent Kernels}

\subsubsection{Definition}

As a linear model after a non-linear embedding, RC has tight links with kernel methods. This class of algorithms implicitly performs a linear regression in the embedding space based on scalar products between pairs of points. Kernels have been studied extensively and, in particular, Random Features have been proposed as finite-dimensional approximations of kernels. 

Similarly, the limit of RC when $N$ goes to infinity is defined as an RK. RC can be interpreted as the temporal equivalent of Random Features: a finite-dimensional approximation of a deterministic RK. 

More precisely, this RK operates on time-dependent scalar products between two reservoir states $\xx\te$ and $\yy\te$ driven by inputs $\ii\te$ and $\jj\te$, respectively. Let $\ww_{r,j}$ and $\ww_{i,j}$ be the $j$-th rows of $\WW_r$ and $\WW_i$ respectively. Eq.~\eqref{eq: reservoir update equation} then yields
\begin{align}
    \langle \xx\tee, \yy\tee \rangle = \frac{1}{N} \sum_j \, & f(\ww_{r,j}^\top \xx\te + \ww_{i,j}^\top \ii\te) \nonumber\\
        &  \times f(\ww_{r,j}^\top \yy\te + \ww_{i,j}^\top \jj\te).
        \label{eq: rc to rk step}
\end{align}
This sum of independent random terms concentrates like Random Features of a certain kernel $k$ \cite{rahimi2008random}:
\begin{equation}
    \langle \xx\tee, \yy\tee \rangle \rightarrow k\left(
    \begin{bmatrix}
        \xx\te\\
        \ii\te
    \end{bmatrix},
    \begin{bmatrix}
        \yy\te\\
        \jj\te
    \end{bmatrix}\right),
    \label{eq: single step kernel limit}
\end{equation}
i.e. a kernel on the concatenation of reservoirs and inputs. This kernel function $k$ is defined by the activation $f$ and the distribution of the random vectors $p(\ww)$. To define deterministic RKs which are not linked to a particular RC algorithm, one needs to remove the dependence in previous embeddings $\xx\te$ and $\yy\te$. This is possible for any rotationally-invariant distribution $p(\ww)$, a more general setting than \cite{dong2020reservoir}. 


To summarize, RKs are deterministic algorithms where the input data $\ii\te$ is used to recurrently update a Gram matrix $\GG\te$, the matrix containing scalar products between all pairs of points. This Gram matrix is initialized arbitrarily $\GG^{(0)}$ and updated as
\begin{equation}
    \GG\tee = k(\GG\te, \{\langle \ii\te, \jj\te\rangle\}_{\ii,\jj}),
    \label{eq: recurrent kernel update}
\end{equation}
where $\{\langle \ii\te, \jj\te\rangle\}_{\ii,\jj}$ denotes the scalar products between all pairs of inputs at time $t$. For conciseness, we use the same letter $k$ for the kernel, a more detailed derivation for common kernels can be found in \cite{dong2020reservoir}. 

Convergence of RC towards its RK limit has been observed in practice in a large range of settings, when the activation function is bounded. Formally, convergence has only been proven in restrictive settings \cite{dong2020reservoir}, which is why convergence will be assessed on a case-by-case basis.

In practice, computing directly with the RK limit is beneficial for medium-sized tasks with a few thousand examples. Iterating them is efficient since it mostly consists of element-wise operations. Kernel methods typically struggle when the number of training points becomes very large as they need to compute scalar products between all pairs of points. 

\subsubsection{Stability}

It is then natural to describe with RKs the limit of this stability metric as $N\rightarrow\infty$. The two reservoirs evolving in parallel define a $2 \times 2$ Gram matrix
\begin{equation}
    \GG_N\te = 
    \begin{pmatrix}
        \left\|\xx_1\te\right\|^2 & \left\langle \xx_1\te, \xx_2\te \right\rangle \\
        \left\langle \xx_1\te, \xx_2\te \right\rangle & \left\|\xx_2\te\right\|^2
    \end{pmatrix}.
    \label{eq: gram matrix definition}
\end{equation}
This matrix is symmetric and invariant by permutation of $\xx_1$ and $\xx_2$. For this reason, we introduce $\Geq\te =  \|\xx_1\te\|^2 =  \|\xx_2\te\|^2$ and $\Gneq\te = \langle \xx_1\te, \xx_2\te \rangle$. 

Since the reservoirs are initialized with independent random draws, the limit at $t=0$ when $N \rightarrow \infty$ is the identity matrix:
\begin{equation}
    \GG^{(0)} = \lim_{N\rightarrow\infty} \GG_N^{(0)} = \II_2.
    \label{eq: limit initial recurrent kernel}
\end{equation}
This defines the initial state of our recurrent kernel. Equivalently, $\Geq^{(0)} = 1$ and $\Gneq^{(0)} = 0$.

We then iterate this recurrent kernel with the input $\ii\te$. 
The stability metric defined for RC in Eq.~\eqref{eq: metric reservoir computing} has a RK equivalent, simply obtained by developing the squared norm,
\begin{equation}
    \LL\te = 2 (\Geq\te - \Gneq\te).
    \label{eq: metric recurrent kernel}
\end{equation}

This simple expression comes from the deterministic nature of RKs. We only have to compute how the two scalar quantities evolve with time instead of distances between high-dimensional vectors. This will enable us to perform analytic studies to quantify the input regularization for example or to tackle the case of discontinuous activation functions.


\subsubsection{Examples}

In this work, we will use three different activation functions, all with Gaussian random weights. The error function $f_1 = \text{erf}$ corresponds to an arcsine kernel in Eq.~\eqref{eq: single step kernel limit}
\begin{equation}
    k_1(\uu, \vv) = \frac{2}{\pi} \arcsin\left( 
        \frac{2 \langle \uu, \vv \rangle}
        {\sqrt{(1+2\|\uu\|^2)(1+2\|\vv\|^2)}}
    \right),
    \label{eq: erf arcsine kernel}
\end{equation}
the sign function $f_2 = \text{sign}$ to
\begin{equation}
    k_2(\uu, \vv) = \frac{2}{\pi} \arcsin\left( 
        \frac{\langle \uu, \vv \rangle}
        {\|\uu\|\|\vv\|}
    \right),
    \label{eq: heaviside arccos kernel}
\end{equation}
and the Rectified Linear Unit $f_3 = \text{ReLU}$ to
\begin{align}
    k_3(\uu, \vv) = \frac{1}{2\pi} &\left(
        \langle \uu, \vv \rangle \arccos\left(-
            \frac{\langle \uu, \vv \rangle}
            {\|\uu\|\|\vv\|}
        \right) \right. \nonumber \\
        & \left. + \sqrt{\|\uu\|^2\|\vv\|^2 - \langle \uu, \vv \rangle^2}
    \right).
    \label{eq: relu arccos kernel}
\end{align}
To link with our case of RKs, $\uu$ and $\vv$ correspond respectively to $\begin{bmatrix}\sigma_r\xx\te \\ \sigma_i\ii\te\end{bmatrix}$ and $\begin{bmatrix}\sigma_r\yy\te \\ \sigma_i\jj\te\end{bmatrix}$ in Eq.~\eqref{eq: single step kernel limit}. These three activation functions have been chosen as they are representative of the diversity between bounded and unbounded functions, as well as Lipschitz-continuous and discontinuous functions. 


\section{Results}

\subsection{Convergence of RC towards RK}

For the classical case of an erf activation function with no input, we know that stability occurs when $\sigma_r < \sqrt{\pi}/2$ \cite{wainrib2016local}. In Fig \ref{fig1: stability}b and \ref{fig1: stability}c, we display the time evolution of the stability metrics $L\te$ and $\mathcal{L}\te$ for $\sigma_r = 0.85 < \sqrt{\pi}/2$ and $\sigma_r = 1.05 > \sqrt{\pi}/2$ respectively. Indeed, we observe that in the first case, the stability metric converges to 0 as $t$ increases, whereas it does not converge to zero in the chaotic regime. 

We see in these results the convergence of RC to the RK limit. As the reservoir size increases, its stability metric $L\te$ tends to the well-defined kernel limit $\mathcal{L}\te$. This motivates our asymptotic stability analysis. In practice, our RK limit accurately describes stability of RC even for reservoirs of moderate sizes around a few hundreds. Finite size effects may appear for smaller sizes and mostly in the chaotic regime. We can thus leverage this deterministic update equation to study stability for various activation functions in the presence of an input. 

In the following, we show exact results for this RK limit. It will describe typical behaviors of RC as well, neglecting finite-size effects.

\subsection{Erf activation function}

\begin{figure*}
    \centering
    \includegraphics[width=\linewidth]{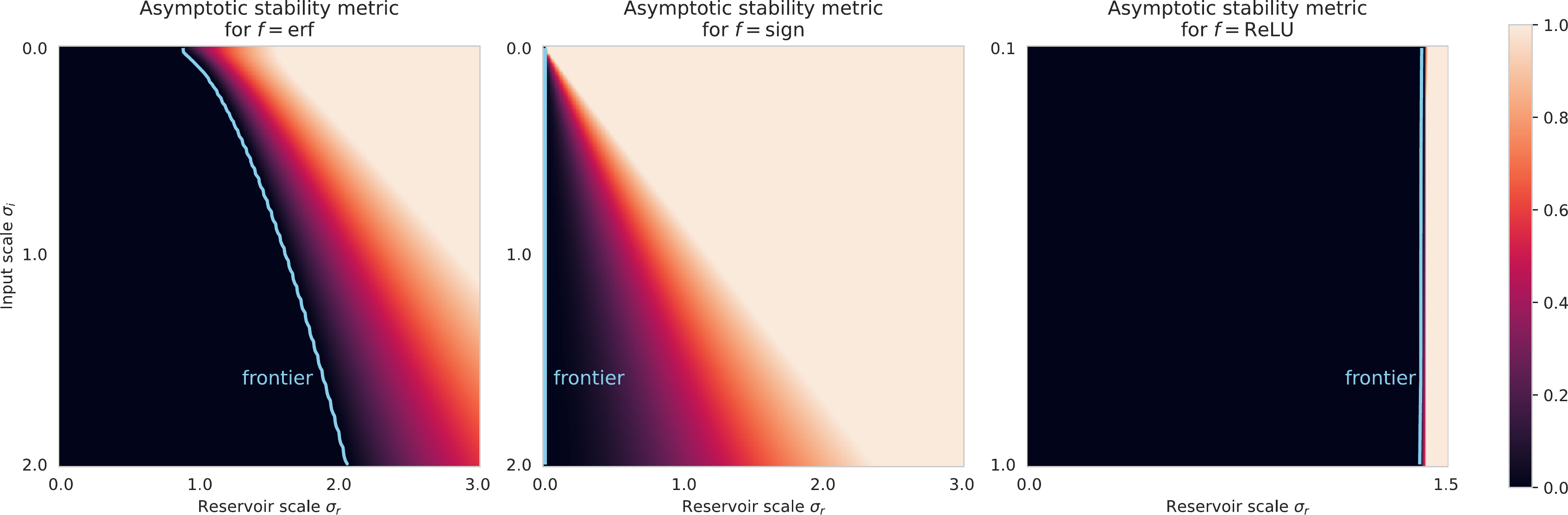}
    \caption{\textbf{(a-c) Asymptotic stability metric for $f = \operatorname{erf}$, $f = \operatorname{sign}$, and $f = \operatorname{ReLU}$ as a function of $\sigma_i$ and $\sigma_r$.} Asymptotic values are computed from the update equations of the recurrent kernel limit, for $t=200$ large enough. When this stability metric converges to 0, the recurrent kernel dynamics are stable. Whenever it converges to a non-zero value, the recurrent kernel dynamics are unstable or chaotic. In blue is drawn the frontier between the stable and chaotic regions.}
    \label{fig2: imshow stability}
\end{figure*}

Eq.~\eqref{eq: metric recurrent kernel} defines two sequences $\Geq\te$ and $\Gneq\te$, initialized with $\Geq^{(0)} = 1$ and $\Gneq^{(0)} = 0$, with update equations deduced from Eq.~\eqref{eq: erf arcsine kernel} for the erf activation function:
\begin{equation}
    \begin{cases}
    \label{eq: Geq Gneq erf update}
    \Geq\tee = \displaystyle\frac{2}{\pi} \arcsin\left(
        \displaystyle\frac{2\sigma_r^2\Geq\te + 2 \sigma_i^2}
        {1+2\sigma_r^2\Geq\te + 2 \sigma_i^2}
    \right) \\
    \Gneq\tee = \displaystyle\frac{2}{\pi} \arcsin\left(
        \displaystyle\frac{2\sigma_r^2\Gneq\te + 2 \sigma_i^2}
        {1+2\sigma_r^2\Geq\te + 2 \sigma_i^2}
    \right)
    \end{cases}
\end{equation}

Thanks to our asymptotic model, we can precisely characterize the transition between stability and chaos. The full derivation is detailed in Appendix~\ref{appB}. 
\begin{theorem}
    \label{thm: erf}
    The two sequences $G_\textnormal{eq}$ and $G_\textnormal{neq}$ are convergent. For any $\sigma_r \geq \sqrt{\pi} / 2$, the frontier between stability and chaos is given by:
    \begin{equation}
        \label{eq: frontier erf thm}
        \psi(\sigma_r) = \frac{4\sigma_r^2}{\pi} - \frac{1}{4} - \frac{2\sigma_r^2}{\pi}\arcsin\left(
            \frac{16\sigma_r^4 - \pi^2}
            {16\sigma_r^4 + \pi^2}
        \right).
    \end{equation}
    \begin{itemize}
        \item If $\sigma_i \geq \psi(\sigma_r)$, the dynamics are stable.
        \item If $\sigma_i < \psi(\sigma_r)$, the dynamics are chaotic.
    \end{itemize}
    Reciprocally, since $\psi$ is a bijection from $[\sqrt{\pi}/2, +\infty)$ to $[0, +\infty)$, for any $\sigma_i \geq 0$:
    \begin{itemize}
        \item If $\sigma_r \leq \psi^{-1}(\sigma_i)$, the dynamics are stable.
        \item If $\sigma_r > \psi^{-1}(\sigma_i)$, the dynamics are chaotic.
    \end{itemize}
\end{theorem}

We display in Fig.~\ref{fig2: imshow stability}a, the limit of $\mathcal{L}\te$ as a function of the hyperparameters $\sigma_r$ and $\sigma_i$. This limit is equal to 0 for small values of $\sigma_r$. This corresponds to the region in which the reservoir is in a stable regime. The limit of $\mathcal{L}\te$ becomes non-zero for large values of $\sigma_r$, which indicates chaotic reservoir dynamics. The frontier between the stable and chaotic regions depends on $\sigma_i$, the standard deviation of the input weights. Having a large input pushes the transition between stability and chaos to larger values of $\sigma_r$.


Without an input, i.e. for $\sigma_i = 0$, we obtain $\psi^{-1}(0) = \sqrt{\pi} / 2$. We thus recover the previous result which is optimal with no input \cite{hermans2012recurrent}. Additionally, we quantify how much the input regularizes the dynamics. This comes from the saturation of the activation function. With a large input, the arguments $u$ of the activations erf$(u)$ are typically larger and we are in the flatter regions of the activation function. The network is therefore less sensitive to changes. 

The quantitative characterization of the frontier may provide a useful tool to restrict the hyperparameter search space. As we want to stay close to this frontier between stability and chaos for optimal performance, it transforms a two-dimensional hyperparameter search on both $(\sigma_i, \sigma_r)$ to a unidimensional search along the frontier. 

\subsection{Sign activation function}


The equations to update the two quantities in Eq.~\eqref{eq: metric recurrent kernel} is deduced from Eq.~\eqref{eq: heaviside arccos kernel} for the sign activation function:
\begin{equation}
    \begin{cases}
    \label{eq: Geq Gneq heaviside update}
    \Geq\tee &= 1
    \\
    \Gneq\tee &= \displaystyle\frac{2}{\pi} \arcsin\left(
        \displaystyle\frac{\sigma_r^2 \Gneq\te + \sigma_i^2}
        {\sigma_r^2+\sigma_i^2}
    \right)
    \end{cases}
\end{equation}

\begin{theorem}
\label{thm: sign}
As soon as $\sigma_r>0$ and for any value $\sigma_i$, the stability metric is converging to a non-zero value, i.e.
\begin{equation}
    \label{eq: sign thm limit}
    \lim_{t\rightarrow\infty} \mathcal{L}\te = l > 0.
\end{equation}
This implies that any reservoir with sign activations is chaotic. 
\end{theorem}

The case of discontinuous activation functions has not received a lot of attention before. In the asymptotic limit, there is an averaging effect that makes the kernel limit continuous. Despite this well-defined limit, there is no rigorous stability in the sense that $\mathcal{L}\te$ converges to 0 exactly. 


Fig.~\ref{fig2: imshow stability}b shows this limit $l$ as a function of $(\sigma_r, \sigma_i)$. We see that there is no stable region, apart from the trivial case $\sigma_r = 0$. However, similar to the saturation effect with erf, the addition of an input seems to regularize the dynamics. For large values of $\sigma_i$, it is possible to still control the stability of the system. More precisely, when $\sigma_i \gg \sigma_r$, we have:
\begin{equation}
    \label{eq: sign taylor}
    l \approx \frac{16 \sigma_r^2}{\pi^2 \sigma_i^2}.
\end{equation}
In this case, the input regularizes the dynamics and $\lim_{\sigma_i \rightarrow \infty} l = 0$. 

Indeed, training with step activation functions has been performed successfully in practice \cite{dong2018scaling,dong2019optical}. This observation may be important for physical implementations of RC or low-power RC with quantized activation functions. 

\subsection{ReLU activation function}


The equations to update the two quantities in Eq.~\eqref{eq: metric recurrent kernel} are deduced from Eq.~\eqref{eq: relu arccos kernel} for the ReLU activation function:
\begin{equation}
    \label{eq: Geq Gneq relu update}
    \begin{cases}
        \Geq\tee =&\, \displaystyle\frac{1}{2}
            \left(\sigma_r^2 \Geq\te + \sigma_i^2\right)
        \\
        \Gneq\tee =&\, \displaystyle\frac{1}{2\pi}
        \left(\sigma_r^2 \Gneq\te + \sigma_i^2\right) \arccos\left(-
            \displaystyle\frac{\sigma_r^2 \Gneq\te + \sigma_i^2}
            {\sigma_r^2 \Geq\te + \sigma_i^2}
        \right) \\
        & + \displaystyle\frac{1}{2\pi} \sqrt{\left(\sigma_r^2 \Geq\te + \sigma_i^2\right)^2 - \left(\sigma_r^2 \Gneq\te + \sigma_i^2\right)^2}
    \end{cases}
\end{equation}

\begin{theorem}
\label{thm: relu}
If $\sigma_r < \sqrt{2}$ and for any value $\sigma_i$, the RK is stable, i.e. we have
\begin{equation}
    \label{eq: relu thm limit}
    \lim_{t\rightarrow\infty} \mathcal{L}\te = 0.
\end{equation}
\end{theorem}

Fig.~\ref{fig2: imshow stability}c shows this limit as a function of $(\sigma_r, \sigma_i)$. We observe that there is no input regularization. In contrast with the erf case, the frontier shows no dependence on $\sigma_i$. This is linked with the absence of saturation in the ReLU activation function. 

For $\sigma_r > \sqrt{2}$, the stability metric diverges and the RK is unstable. An interesting point to notice is that despite ReLU being 1-Lipschitz, the frontier is not for $\sigma_r = 1$ as it could have been predicted from classical analysis of the Lipschitz constant. Instead, slightly larger reservoir weights are possible, thanks to the subdifferentiability of ReLU at zero. 

To apply these results to RC, particular care needs to be taken here regarding the convergence of RC towards its RK limit. This convergence is quite robust in practice for bounded activation functions but it is not always the case with ReLU activations. Indeed, we show in Appendix \ref{app: rc convergence} that convergence is obtained in a large part of the stable region but not in the unstable region. 

\section{Discussion}

In this work, we have presented a framework to study the asymptotic stability of RC. We relied on the recurrent kernel limit to quantitatively characterize trajectories when the reservoir size is large. We then applied our framework to three different activation functions. We showed the importance of having a continuous activation function and made the link between input regularization and saturation of the activation function.

These results can be important in practice for hyperparameter tuning. They also help to develop a better understanding of stability in non-classical cases. We believe this framework is powerful enough to be applied to a wide range of applications. 

In the future, more general results with strong convergence proofs of RC towards RKs may be derived, supporting the observations presented here. This study may be generalized to a larger class of functions, like the hyperbolic tangent which is commonly used in RC. The observed behaviors could be extended to any differentiable and saturating activation function. The corresponding kernel and related quantities may not have an analytic expression, but they can still be computed with integrals. 

One may also extend this approach to other RC architectures such as Deep Reservoir Computing \cite{gallicchio2017deep, gallicchio2020deep}. As a more general comment, this kernel approach may be relevant for non-recurrent architectures as well, to understand better the propagation of perturbations in neural networks. 

The associated code is available at \cite{coderepository}. 

\section*{Acknowledgements}

We would like to thank Pakshal Bohra and Tony Wu for their insightful comments to revise this paper. 

\section*{Funding}

J.D. and M.U. acknowledge funding from European Research Council (ERC) under the European Union’s Horizon 2020 research and innovation programme (Grant Agreement No. 101020573 FunLearn). 

\printbibliography

\appendices

\section{Convergence of RC towards RK}
\label{app: rc convergence}

\begin{figure*}
    \centering
    \includegraphics[width=\linewidth]{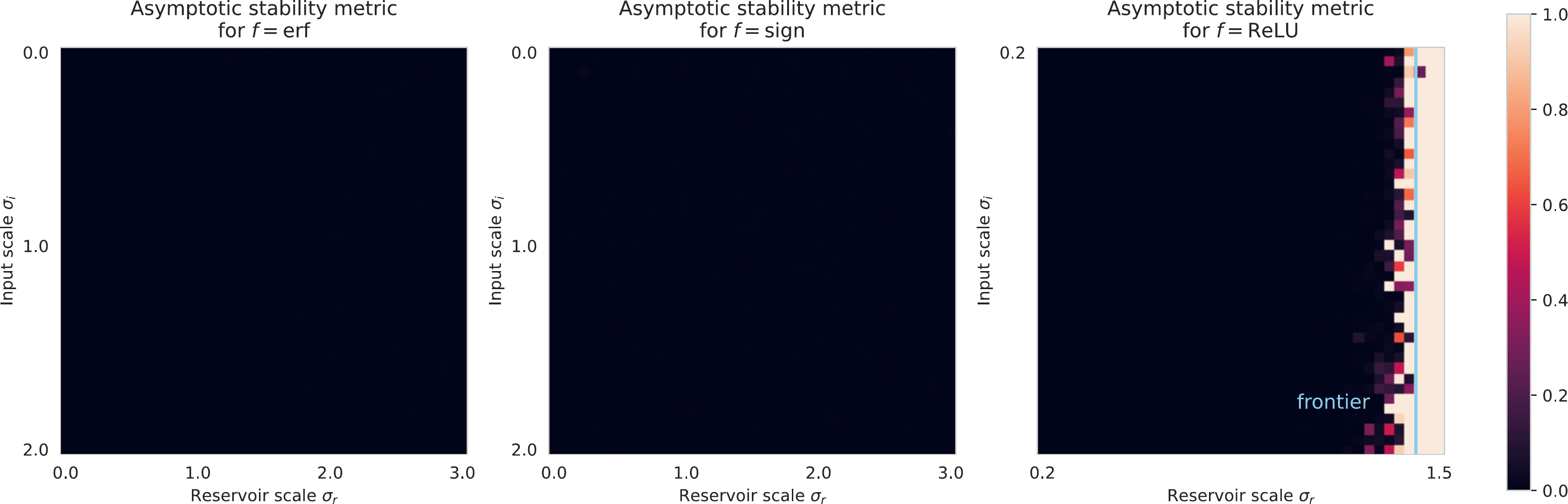}
    \caption{\textbf{(a-c) Convergence study of RC towards RK for $f = \operatorname{erf}$, $f = \operatorname{sign}$, and $f = \operatorname{ReLU}$ as a function of $\sigma_i$ and $\sigma_r$.} Frobenius norm $E$ between the final Gram matrices obtained from a reservoir of size $N=2000$ and the Recurrent Kernel equivalent, iterated with two different random inputs $\ii\te$ and $\jj\te$ of length $50$. We observe robust convergence when the activation function is bounded or for small values of $\sigma_r$.}
    \label{fig3: convergence}
\end{figure*}

The results shown in Fig.~\ref{fig2: imshow stability} are exact for RKs, but convergence of RC towards the RK limit is required to translate them in RC. This convergence needs to be assessed for each combination of hyperparameters $(\sigma_i, \sigma_r)$. 

We evaluate this convergence by iterating both a reservoir of size $N = 2000$ and the associated RK for each activation function presented previously. They are fed two random input time series $\ii\te$ and $\jj\te$ of length $50$ and the final $2 \times 2$ Gram matrices are computed, denoted $G$ for RC and $\mathcal{G}$ for the associated RK. Our convergence metric is defined as
\begin{equation}
    \label{eq: convergence metric}
    E = \|G-\mathcal{G}\|^2_F
\end{equation}
with $\|\cdot\|_F$ the Frobenius norm.

We see in Fig.~\ref{fig3: convergence} that convergence is reliably obtained with erf and sign activation functions. On the other hand, with an unbounded ReLU activation function, convergence does not happen for large values of $\sigma_r$. This implies that the previous theorems also hold for RC, apart from the ReLU case for large $\sigma_r$. This apparent link between stability and convergence of RC towards a kernel limit calls for more investigation.

\section{Technical results for $f = \operatorname{erf}$}
\label{appB}

\subsection{Study of $\Geq$}

We define the function $h_1$ on $[0, 1]$ by
\begin{equation}
    h_1 : x \mapsto \frac{2}{\pi} \arcsin\left(
        \frac{2\sigma_r^2 x + 2\sigma_i^2}{1 + 2\sigma_r^2 x + 2\sigma_i^2}
    \right)
\end{equation}
such that $\Geq\tee = h_1\left(\Geq\te\right)$. 

Since $h_1(0) = \rarcsin\left(\frac{2 \sigma_i^2}{1+2\sigma_i^2}\right) > 0$, $h_1(1) = \rarcsin\left(1-\frac{1}{1+2\sigma_r^2+2\sigma_i^2}\right) < 1$, and the continuity of $h_1$, the intermediate value theorem ensures the existence of at least one fixed point for $h_1$.

Moreover, $h_1$ is strongly concave, as it is twice differentiable with
\begin{equation}
    h_1''(x) = - \frac{16 \sigma_r^4 (1+3\sigma_r^2x+3\sigma_i^2)}{\pi (1+2\sigma_r^2x+2\sigma_i^2)^2 (1+4\sigma_r^2x+4\sigma_i^2)^{3/2}} < 0
\end{equation}
for $x \in [0, 1]$. $h_1$ thus has at most two fixed points.

If $h_1$ had two fixed points $a_1 < a_2$, then the strong concavity of $h_1$ would imply $h_1(0) < a + \frac{b-a}{b-a} (0-a) = 0$, which contradicts the first observation of this proof. Thus, $h_1$ has a unique fixed point that we denote by $a$.

Since $h_1(1) < 1$, $h_1(x) \neq x$ for $x \in (a, 1]$, and because $h_1$ is continuous, we necessarily have $h_1(x) < x$ for all $x \in (a, 1]$. The sequence $\Geq$ is thus decreasing. As it is non-negative, thus bounded below, it converges to $a$, the fixed point of $h_1$. 

\subsection{Study of $\Gneq$}

We define the function $h_2^g$, defined on $[0, a]$ and parametrized by $g \in [a, 1]$, by
\begin{equation}
    h_2^g : x \mapsto \frac{2}{\pi} \arcsin\left(
        \frac{2\sigma_r^2 x + 2\sigma_i^2}{1 + 2\sigma_r^2 g + 2\sigma_i^2}
    \right)
\end{equation}
such that $\Gneq\tee = h_2^{\Geq\te}\left(\Gneq\te\right)$. As $\Geq$ converges to $a$, we will study the sequence $\gneq$ defined by $\gneq^{(0)} = 0$ and $\gneq\tee = h_2^a\left(\gneq\te\right)$. 

Since $h_2^a$ is strongly convex as it is of the form $h_2^a(x) = A \arcsin(B x + C)$ with $A, B, C > 0$, it has at most two fixed points. One of them is $a$, since 
\begin{equation}
    h_2^a(a) = \frac{2}{\pi} \arcsin\left(
        \frac{2\sigma_r^2 a + 2\sigma_i^2}{1 + 2\sigma_r^2 a + 2\sigma_i^2}
    \right) = h_1(a) = a.
\end{equation}

Let $b$ be the smallest fixed point of $h_2^a$. Because $h_2^a$ is an increasing non-negative function, $0 \leq h_2^a(x) \leq b$ and $\gneq$ stays in $[0, b]$. 

$h_2^a(x) \neq x$ for $x < b$ by definition of $b$. As $h_2^a(0) > 0$ and using the continuity of $h_2^a$, $h_2^a(x) > x$ for $x \in [0, b)$. $\gneq$ is hence an increasing sequence. Because it is bounded above, it converges to $b$, the unique fixed point of $h_2^a$ restricted to $[0, a]$.

\subsection{Equation of the frontier}

Using Eq. \eqref{eq: metric recurrent kernel}, the limit of $\mathcal{L}\te$ is
\begin{equation}
    \lim_{t \rightarrow \infty} \mathcal{L}\te = 2 (a-b).
\end{equation}
We want to study when $a$ is the smallest fixed point of $h_2^a$.

This property is linked with the derivative $(h_2^a)'(a)$. Since $h_2^a$ is strongly convex with $h_2^a(0) > 0$, if $(h_2^a)'(a) > 1$, there exists another fixed point in $(0, a)$, while when $(h_2^a)'(a) < 1$, $a$ is the smallest fixed point of $h_2^a$. The derivative is given by
\begin{equation}
    (h_2^a)'(g) = \frac{4\sigma_r^2}{\pi\sqrt{1 + 4\sigma_r^2 g + 4 \sigma_i^2}}.
\end{equation}

The frontier corresponds to the equation $(h_2^a)'(a) = 1$. This equation can be rewritten as:
\begin{equation}
    \label{eq: erf thm proof sigma_i}
    a = \frac{4 \sigma_r^2}{\pi^2} - \frac{1}{4\sigma_r^2} - \frac{\sigma_i^2}{\sigma_r^2}.
\end{equation}

Injecting this in the equation defining $a$ as a fixed point of $h_1$, i.e.
\begin{equation}
    a = \rarcsin\left(
        \frac{2\sigma_r^2 a + 2 \sigma_i^2}{1+2\sigma_r^2 a + 2 \sigma_i^2}
    \right),
\end{equation}
yields the desired equation for the frontier. 


\section{Technical results for $f = \operatorname{sign}$}
\label{app: sign}

We define the function $h_2$ on $(0, 1)$ by
\begin{equation}
    h_2 : x \mapsto \rarcsin\left(
        1 - \frac{\sigma_r^2 (1-x)}{\sigma_r^2 + \sigma_i^2}
    \right),
\end{equation}
such that $\Gneq\tee = h_2\left(\Gneq\te\right)$. $\Gneq$ corresponds to fixed point iteration of $h_2$, starting at $\Gneq^{(0)} = 0$. It therefore converges to the smallest fixed point of $h_2$.

$h_2$ is strictly convex, and thus has at most two fixed points. $x=1$ corresponds to one such fixed point with a vertical tangent. Since $h_2(0) > 0$, there is another fixed point in $(0, 1)$, that we denote by $b$. In the end, $\lim_{t \mapsto \infty} \mathcal{L}\te = 2 (1 - b) > 0$.

For $\sigma_i \gg \sigma_r$, the asymptotic approximation of $\arcsin$ gives
\begin{align}
    1 - b &= 1 - h_2(b) \\
    &= 1 - \rarcsin\left(
        1 - \frac{\sigma_r^2 (1-b)}{\sigma_i^2} + O\left(\frac{\sigma_r^4}{\sigma_i^4}\right)
    \right) \\
    &= 1 - \frac{2}{\pi} \left(
        \frac{\pi}{2} - \sqrt{2} \sqrt{\frac{\sigma_r^2 (1-b)}{\sigma_i^2}} 
    \right) + O\left(\frac{\sigma_r^3}{\sigma_i^3}\right) \\
    &= \frac{2\sqrt{2}\sigma_r\sqrt{1-b}}{\pi \sigma_i} + O\left(\frac{\sigma_r^3}{\sigma_i^3}\right).
\end{align}

Taking the square, we finally obtain
\begin{equation}
    1-b = \frac{8 \sigma_r^2}{\pi^2 \sigma_i^2} + O\left(\frac{\sigma_r^3}{\sigma_i^3}\right).
\end{equation}

\section{Technical results for $f = \operatorname{ReLU}$}
\label{app: relu}

We define the function $h_1$ on $\mathbb{R}_+$ by
\begin{equation}
    h_1 : x \mapsto \frac12 \left(\sigma_r^2 x + \sigma_i^2\right).
\end{equation}
We have $\Geq\tee = h_1\left(\Geq\te\right)$. Since $h_1$ is an affine function, the study of its fixed points is straightforward. 

When $\sigma_r < \sqrt{2}$, $h_1$ has a unique fixed point $a$ given by
\begin{equation}
    a = \frac{\sigma_i^2}{2-\sigma_r^2}.
\end{equation}
Since $h_1(x) \geq x$ for $x \leq a$, $\Geq$ is an increasing sequence, and therefore converges to $a$. 

We define the function $h_2$ on $[0, a]$ by
\begin{align}
    h_2^g : x \mapsto & \displaystyle\frac{1}{2\pi}
        \left(\sigma_r^2 x + \sigma_i^2\right) \arccos\left(-
            \displaystyle\frac{\sigma_r^2 x + \sigma_i^2}
            {\sigma_r^2 g + \sigma_i^2}
        \right) \nonumber\\
        & + \displaystyle\frac{1}{2\pi} \sqrt{\left(\sigma_r^2 g + \sigma_i^2\right)^2 - \left(\sigma_r^2 x + \sigma_i^2\right)^2}.
\end{align}
We have $\Gneq\tee = h_2^{\Geq\te}\left(\Gneq\te\right)$. As $\Geq$ converges to $a$, we will study the sequence $\gneq$ defined by $\gneq^{(0)} = 0$ and $\gneq\tee = h_2^a\left(\gneq\te\right)$. 

First of all, $a$ is a fixed point of $h_2^a$. We then compute the first derivative:
\begin{equation}
    (h_2^a)'(x) = \frac{1}{2\pi} \sigma_r^2 \arccos\left(-
        \displaystyle\frac{\sigma_r^2 x + \sigma_i^2}
        {\sigma_r^2 a + \sigma_i^2}
    \right).
\end{equation}
In particular, $(h_2^a)'(a) = \sigma_r^2 / 2 < 1$. 

Moreover the second derivative is
\begin{equation}
    (h_2^a)''(x) = \frac{\sigma_r^4}{2 \pi \sqrt{(\sigma_r^2a+\sigma_i^2)^2 - (\sigma_r^2x+\sigma_i^2)^2}} > 0
\end{equation}
for all $x \in [0, a)$. Thanks to these two inequalities, $a$ is the unique fixed point of $h_2^a$ in $[0, a]$, and $h_2^a(x) > x$ for $x \in [0, a)$. $\gneq$ is an increasing sequence bounded above, thus converges towards $a$, the fixed point of $h_2^a$. 

\end{document}